\def\year{2022}\relax
\documentclass[letterpaper]{article} 
\usepackage{aaai22}  
\usepackage{times}  
\usepackage{helvet}  
\usepackage{courier}  
\usepackage[hyphens]{url}  
\usepackage{graphicx} 
\usepackage{natbib}  
\usepackage{caption} 
\DeclareCaptionStyle{ruled}{labelfont=normalfont,labelsep=colon,strut=off} 
\usepackage{algorithm}
\usepackage{algorithmic}
\usepackage{caption}
\usepackage{subcaption}
\usepackage{multirow}
\usepackage{amsmath,amssymb}
\usepackage{booktabs}
\usepackage{xcolor,colortbl}
\usepackage[export]{adjustbox}

\usepackage{arydshln}

\usepackage{newfloat}
\usepackage{listings}
\floatstyle{ruled}
\newfloat{listing}{tb}{lst}{}
\floatname{listing}{Listing}

\definecolor{yellow}{rgb}{1.0,0.93,0.63}
\definecolor{dark-red}{rgb}{1.0,0.93,0.63}
\definecolor{dark-blue}{rgb}{1.0,0.93,0.63}
\definecolor{dark-orange}{rgb}{1.0,0.93,0.63}
\definecolor{dark-purple}{rgb}{1.0,0.93,0.63}
\newcommand{\better}{\cellcolor{yellow}}
\nocopyright 

\title{IMB-NAS: Neural Architecture Search for Imbalanced Datasets}
\author{
    Rahul Duggal\textsuperscript{\rm 1},
    Sheng-Yun Peng\textsuperscript{\rm 1},
    Hao Zhou,
    Duen Horng Chau\textsuperscript{\rm 1}
}
\affiliations{
\textsuperscript{\rm 1} Georgia Institute of Technology\\
\{rahulduggal, shengyun, polo\}@gatech.edu, zhhoper@gmail.com

}

\usepackage{bibentry}

\begin{document}

\maketitle

\begin{abstract}
    Class imbalance is a ubiquitous phenomenon occurring in real world data distributions. 
  To overcome its detrimental effect on training accurate classifiers, existing work follows three major directions: class re-balancing, information transfer, and representation learning.
    In this paper, we propose a new and complementary direction for improving performance on long tailed datasets---optimising the \textit{backbone architecture} through neural architecture search (NAS).
  We find that an architecture's accuracy obtained on a balanced dataset is not indicative of good performance on imbalanced ones.
  This poses the need for a full NAS run on long tailed datasets which can quickly become prohibitively compute intensive.
To alleviate this compute burden, we aim to efficiently adapt a NAS super-network from a balanced source dataset to an imbalanced target one.
  Among several adaptation strategies, we find that the most effective one is to retrain the linear classification head with reweighted loss, while freezing the backbone NAS super-network trained on a balanced source dataset.
  We perform extensive experiments on multiple datasets and provide concrete insights to optimise architectures for long tailed datasets.

\end{abstract}

\section{Introduction}

The natural world follows a long tail data distribution wherein a small percentage of classes constitute the bulk of data samples, while a small percentage of data is distributed across numerous minority classes.
Training accurate classifiers on imbalanced datasets has been an active research direction since the early 90s. 
Much of prior work~\cite{kang2019decoupling,zhou2020bbn, duggal2021har} centers on improving the performance (measured via accuracy) of a fixed backbone architecture such as ResNet-32.
In this work, we take a complementary direction and aim to optimise the backbone architecture via neural architecture search.
Indeed this is an important direction since prevalent practices demand that neural architectures be optimised to fit the size/latency constraints of tiny edge devices.

\begin{table}
        \centering
        \begin{tabular}{cccccc}
        \toprule
        Dataset & Model & Flops   & \multicolumn{2}{c}{Accuracy (\%)}                \\
                &       &         & bal($1\times$) & imbal($100\times$) \\
        \midrule 
        \multirow{2}{*}{Cifar10} & A1 & 410  & {94.6} & {77.3}\\
        & A2 & 407  & {94.7} & {74.1}\\ 
        \cdashline{1-6}[.7pt/1.5pt]\noalign{\vskip 0.15em}
        \multirow{2}{*}{Cifar100} & A3 & 400 & {76.1} & {39.4}\\
        & A4 & 179 & {75.0} & {43.0}\\ \bottomrule
       \end{tabular}
     \caption{\textbf{Motivation.} We sample four architectures A1-A4 from the DARTS search space and train them on balanced (i.e. $1\times$) and imbalanced versions (i.e. $100\times$) of Cifar10 and Cifar100. (\textbf{Top}) Two similarly sized architectures (A1,A2) achieve similar accuracy on balanced Cifar10, but differ by $3\%$ in presence of $100\times$ imbalance. (\textbf{Bottom}) The larger architecture (A3) outperforms the smaller on (A4) on balanced Cifar100, but under performs by $3.6\%$ in the presence of $100\times$ imbalance. This suggests that an architecture's performance on balanced datasets is not indicative of it's performance on imbalanced ones.}
     \label{tab:crown_jewel}
\end{table}

To optimise the backbone architecture, we rely on the recent work from Neural Architecture Search (NAS)~\cite{guo2020single} that optimises a neural network's  architecture primarily on datasets that are \textit{balanced} across classes. 
This workflow naturally prompts the question: is the architecture optimised on a class balanced dataset  also the optimal one for imbalanced datasets?
Table~\ref{tab:crown_jewel} provides evidence to the contrary.
The first row shows two architectures--A1,A2--sampled from the DARTS search space~\cite{liu2018darts} having similar size and accuracy on balanced Cifar10, but an accuracy gap of $3\%$ in the presence of $100\times$ imbalance.
The second row compares a larger architecture A3 outperforms a smaller one A4 on balanced Cifar100.
However, in the presence of $100\times$ imbalance, the smaller architecture outperforms the larger one by more than $3\%$.
These results and more in Sec~\ref{subsec:rank_transfer}, indicate that the optimal architecture on a balanced dataset may not be the optimal one for imbalanced datasets.
This means each target imbalanced dataset requires its own NAS procedure to obtain the optimal architecture. 
%

Running a NAS procedure for each target dataset is computationally expensive and quickly becomes intractable in the presence of multiple target datasets. 
To overcome the compute burden of running NAS from scratch, we formalize the task of architectural rank adaptation from balanced to imbalanced datasets.
Towards this task, Section~\ref{subsec:rank_adjustment} describes two intuitive rank adaptation procedures that either fine-tune the classifier only, or together with the backbone.
%
%
Our comprehensive experiments reveal the key insight that the adaptation procedure is most affected by the linear classification head trained on top of the backbone. 
Armed with this insight, we propose to re-use a NAS super-net backbone trained on balanced data and re-train only the classification head to efficiently adapt a pre-trained NAS super-net for imbalanced data.
This is extremely efficient since it involves training only a linear layer on top of the pre-trained super-network.
%

Overall, our contributions in this work are: 
\begin{enumerate}
    \item \textbf{New insight.}  We show that architectural rankings transfers poorly from balanced to imbalanced datasets. 
    
    \item \textbf{Novel task.} We construct the novel task to efficient adapt a NAS super-network from balanced to imbalanced datasets. 
    
    \item \textbf{Novel solution.} We propose a simple and efficient solution--retraining the classifier head while freezing the backbone--to efficiently adapt a NAS super-network from balanced to imbalanced datasets.
    
    
    
    
\end{enumerate}

\section{Related Works}
We cover relevant work from three related areas.
\subsection{Overcoming long tail class imbalance} Prior work on tackling long tail imbalance can divided into three broad areas (see survey~\cite{zhang2021deep}): class-rebalancing that includes data re-sampling (SMOTE~\cite{chawla2002smote}, ADASYN~\cite{he2008adasyn}), loss re-weighting~\cite{kang2019decoupling,cui2019class,duggal2020elf,duggal2021har}, logit adjustment~\cite{menon2020long,tian2020posterior,zhang2021distribution}; 
Information augmentation that includes transfer learning~\cite{wang2017learning,yin2019feature}, data augmentation~\cite{chu2020feature}; and
module improvement that encompasses methods in representation learning~\cite{liu2019large}, classifier design~\cite{wu2020solving}, decoupled training~\cite{kang2019decoupling} and ensembling~\cite{zhou2020bbn}.
Different from all of the existing works, our work explores a new direction of performance improvement on long tail datasets--that via optimising the backbone architecture. 
This complements existing approaches and can work in tandem to further boost accuracy and efficiency on imbalanced datasets.

\subsection{Neural architecture search}
Prior work on architecture search can be categorized in improving its three main pillars (see survey~\cite{elsken2019neural})---Search space design with the idea of incorporating a large diversity of architectures.
Popular spaces include cell based spaces such as NASNets~\cite{zoph2018learning}, and recent spaces from the ShuffleNet~\cite{zhang2018shufflenet} and MobileNet~\cite{howard2017mobilenets} model families. 
The second pillar constitutes search strategy design to efficiently locate performant architectures from the search space.
Popular strategies involve reinforcement learning~\cite{baker2016designing,zoph2018learning}, evolutionary algorithms~\cite{real2017large,duggal2021compatibility} or gradient descent on continuous relaxations of the search space~\cite{liu2018darts}.
The third pillar constitutes performance estimation strategies~\cite{baker2017accelerating,falkner2018bohb} with the goal of cheaply estimating the goodness (in terms of accuracy or efficiency) of an architecture.
All of the above works search optimal architectures on datasets that are fully balanced across all classes.
Our experiments however show that the set of optimal architectures differ significantly from balanced to imbalanced datasets.
This calls for developing new NAS methods or efficient adaptation strategies (e.g. this work) to search for optimal architectures on real world, imbalanced datasets. 

\subsection{Architecture transfer}
We summarize prior work on evaluating robustness of architectures to distributional shifts in the training dataset. 
Neural Architecture Transfer~\cite{lu2021neural} explore architectural transferability from large-scale to small-scale fine grained datasets. 
%
%
However, there are two limitations--the source and target datasets considered in this work are balanced across all classes and additionally  this work assumes all target datasets are known apriori which is infeasible in many industry use-cases.
NASTransfer~\cite{panda2021nastransfer} consider transferability between large-scale imbalanced datasets including ImageNet-22k which is a highly imbalanced dataset.
%
%
%
Their approach is practically useful for very large datasets (e.g. ImageNet-22k) for whom direct search is prohibitive, however when it is feasible (e.g. on ImageNet) direct search typically leads to better architectures than proxy search. 
Differing from these, our work advocates to directly adapt a super-network pre-trained on fully balanced datasets (instead of proxies) to imbalanced ones.
A key feature of such adaptation is efficiency---the compute required for such an adaptation needs to be much lesser than that for repeating the search on the target dataset.
\section{Methodology}
\subsection{Notation}
Assume $\mathcal{D} = \{x_1,y_i\}$ denotes the training dataset of images where $y_i$ is the label for image $x_i$. Let $n_j$ specify the number of training images in class $j$. After sorting the classes by cardinality in decreasing order, the long tail assumption specifies that if $i<j$, then $n_i \geq n_j$ and $n_1>>n_C$. We use $\phi$ to denote a deep neural network that is composed of a backbone $\phi(a,w_a)$ with architecture $a$, weights $w_a$ and a linear classifier $\phi(w_c)$. The model $\phi$ is trained using a training loss and loss re-weighting strategy. On balanced datasets, we use the cross entropy loss (denoted as CE) to train a neural network. For imbalanced datasetsm we additionally incorporate the effective re-weighting strategy~\cite{cui2019class} that reweights samples from class $j$ with $\frac{1-\beta}{1-\beta^{n_j}}$ where $\beta$ is a hyperparameter. Following previous works~\cite{cao2019learning,duggal2020elf}, the re-weighting strategy is applied after a delay of few training epochs which is denoted using the shorthand DRW. 


\begin{figure}[!h]%
    \includegraphics[width=\linewidth]{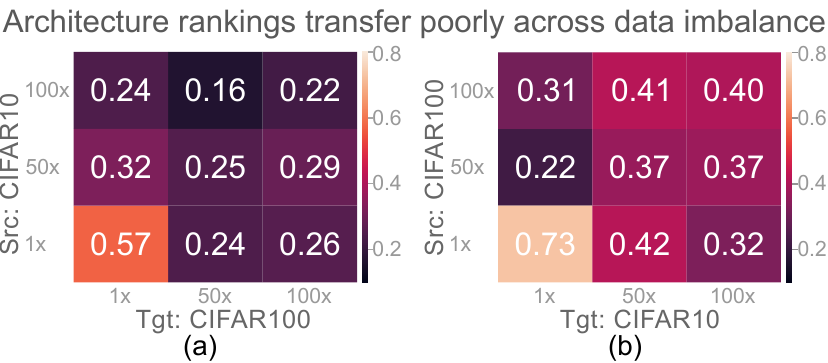}
    \caption{\textbf{Evaluating architectural transferability.} We train all 149 Mflop architectures from NATS-Bench on Cifar10, Cifar100 with $1\times, 50\times, 100\times$ imbalance and compute kendall tau correlation between the rank orderings on all datasets. We observe high correlation (bottom left cells) when both $\mathcal{D}_s, \mathcal{D}_t$ are balanced, and low correlation otherwise. This means that architectural rankings transfer poorly across data imbalance.}%
    \label{fig:correlation_exp}%
\end{figure}

\subsection{Architecture ranking transfer: A motivating experiment}
\label{subsec:rank_transfer}
We study the impact of backbone architecture on imbalanced datasets using the following experiment.
We construct an architecture search space $\mathcal{A}$ by sampling all 149 Mega Flops architectures from the NATS-bench search space~\cite{dong2021nats}.
Overall $\mathcal{A}$ contains 135 architectures with exactly the same learning capacity (or Flops), but different architectural patterns (e.g. kernel sizes, layer connectivity).
The architectures in $\mathcal{A}$ are trained on the source and target datasets $D_s, D_t$ using loss function CE on balanced datasets, and the re-weighted loss function CE+DRW on imbalanced ones.
Following this, the architectures are ranked based on validation accuracy and the kendall Tau metric is computed between the rank orderings obtained on $D_s$ and  $D_t$.
%
A high correlation means similar architectural rankings on both datasets, while a low correlation implies widely different rankings.

Figure~\ref{fig:correlation_exp} presents the outcomes on two scenarios: (1) $D_s$ is Cifar10 at three levels of imbalance ($1\times, 50\times, 100\times)$ and $D_t$ is Cifar100 at the same imbalance levels; and (2) the opposite direction.
%
%
There are two major observations---First, the high correlation in the bottom left square indicates that the architectural rankings transfer quite well across balanced datasets.
Second, the low correlation for all other cells indicates low transferability across imbalanced datasets.
This means the rank orderings on imbalanced datasets widely differs from that on balanced ones.

To avoid the compute burden of performing a NAS run on every target imbalanced dataset, we develop efficient ``adaptation'' procedures to adapt a NAS super-net from balanced to imabalanced datsets.
Before going into the details, in the next section we provide a brief overview of exisiting NAS methods.

%
%

\subsection{Revisiting neural architecture search}
We look at sampling based NAS methods that involve two steps. The first step involves training a super-network with backbone $\phi(a, w_a)$ and classifier $\phi(w_c)$ on a training dataset $D$ via the following minimization
\begin{equation}
    w^{*}_{a, \mathcal{D}}, w^{*}_{c,  \mathcal{D}} = \mathop{min}_{w_a,w_c} \mathop{\mathbb{E}}_{a \sim \mathcal{A}} \left( \mathcal{L}(\phi(w_c),\phi(a, w_a); \mathcal{D})   \right).
\end{equation}
Here the inner expectation is performed by sampling architectures $a$ from a search space $\mathcal{A}$ via uniform, or attentive sampling.

The second step involves searching the optimal architecture that maximizes validation accuracy via the following optimisation
\begin{equation}
    a^*_{\mathcal{D}} = \mathop{max}_{a \in \mathcal{A}} Acc \left( \phi(w_c),\phi(a, w_a); \mathcal{D})   \right).
\end{equation}
This maximization is typically implemented via evolutionary search or reinforcement learning.
Next, we discuss efficient adaptation procedures to adapt a NAS super-net trained on a balanced dataset onto an imbalanced one.

\subsection{Rank adaptation procedures}
\label{subsec:rank_adjustment}
Given source and target datasets $\mathcal{D}_s, \mathcal{D}_t$, we first train a super-network on $D_s$ by solving the following optimisation
\begin{equation}
\label{eq:p0_eqn}
    w^{*}_{a, \mathcal{D}_s}, w^{*}_{c,  \mathcal{D}_s} = \mathop{min}_{w_a,w_c} \mathop{\mathbb{E}}_{a \sim \mathcal{A}} \left( \mathcal{L}(\phi(w_c),\phi(a, w_a); \mathcal{D}_s)   \right).
\end{equation}
Our goal then is to adapt the optimal super-net weights $w^{*}_{a, \mathcal{D}_s}, w^{*}_{c,  \mathcal{D}_s}$ found on $D_s$ to the target dataset $D_t$ which suffers from class imbalance. The most efficient adaptation procedure involves freezing the backbone, while adapting only the linear classifier on $D_t$ by minimizing the re-weighted loss $\mathcal{L}_{RW}$
\begin{equation}
\label{eq:p1_eqn}
    w^{*}_{c,  \mathcal{D}_t} = \mathop{min}_{w_c} \mathop{\mathbb{E}}_{a \sim \mathcal{A}} \left( \mathcal{L}_{RW}(\phi(w_c),\phi(a, w^{*}_{a, \mathcal{D}_s}); \mathcal{D}_t)   \right).
\end{equation}
The resulting super-network contains backbone weights $w^{*}_{a, \mathcal{D}_s}$ trained on $\mathcal{D}_s$ and classifier weights $w^{*}_{c,  \mathcal{D}_s}$ trained on $\mathcal{D}_t$.  Solving the above optimisation is extremely efficient since most of the network is frozen while only the classifier is trained. On the other hand, one could also adapt the backbone by fine-tuning on the target dataset. This is achieved by minimizing the delayed re-weighted loss $\mathcal{L}_{DRW}$
\begin{equation}
\label{eq:p2_eqn}
    w^{**}_{a,  \mathcal{D}_t}, w^{*}_{c,  \mathcal{D}_t} = \mathop{min}_{w_a, w_c} \mathop{\mathbb{E}}_{a \sim \mathcal{A}} \left( \mathcal{L}_{DRW}(\phi(w_c),\phi(a, w^{*}_{a, \mathcal{D}_s}); \mathcal{D}_t)   \right).
\end{equation}
 Here, the double star on $w^{**}_{a,  \mathcal{D}_t}$ indicates the weights were obtained via fine-tuning $w^{*}_{a, \mathcal{D}_s}$ using one tenth of the original learning rate and one third the number of original training epochs. Also, recall that the delayed re-weighted loss $\mathcal{L}_{DRW}$ is nothing but the unweighted loss $\mathcal{L}$ in the first few epochs and the re-weighted loss $\mathcal{L}_{RW}$ subsequently. Note that our second adaptation procedure is more compute intensive since the backbone is also adapted, but still much less intensive than running the full search on the target dataset.
 
Our final and most compute intensive procedure involves directly searching on the target dataset via $\mathcal{L}_{DRW}$. This is achieved via the following minimization 
 \begin{equation}
 \label{eq:p3_eqn}
    w^{*}_{a, \mathcal{D}_t}, w^{*}_{c,  \mathcal{D}_t} = \mathop{min}_{w_a,w_c} \mathop{\mathbb{E}}_{a \sim \mathcal{A}} \left( \mathcal{L}_{DRW}(\phi(w_c),\phi(a, w_a); \mathcal{D}_t)   \right).
\end{equation}

\begin{table}[]
\begin{tabular}{ccl}
\toprule
Adj & Eqn & Description                             \\ \midrule
P0           &      (\ref{eq:p0_eqn})        & No adaptation. \\
P1           &      (\ref{eq:p1_eqn})        & Freeze backbone, retrain classifier on $\mathcal{D}_t$.                      \\
P2           &      (\ref{eq:p2_eqn})        & Finetune backbone and retrain classifier on $\mathcal{D}_t$      \\
P3           &      (\ref{eq:p3_eqn})        & Re-train backbone and classifier on $\mathcal{D}_t$.\\ \bottomrule
\end{tabular}
\caption{Summarizing rank adaptation procedures.}
\label{tab:rank_adjustment_summary}
\end{table}

The three adaptation procedures and their associated compute costs are summarized in Table~\ref{tab:rank_adjustment_summary}.

\section{Experiments}
We begin this section by answering which rank adaptation procedure works best, both in terms of efficiency of the procedure and the accuracy of the resulting networks. We then 
perform an extensive ablation study to uncover the effect of different design choices. 

\subsection{Implementation details}
We implement our methods using Pytorch on a system containing 8 V100 GPUs.
Other details are as follows:

\smallskip
\noindent \textbf{Datasets.} We construct imbalanced versions of Cifar-10 and Cifar-100 by sub-sampling from their original training splits~\cite{cui2019class}. The c$^{th}$ class in the resulting datasets contains $n_c = n \mu^{c}$ examples where $n$ is the original cardinality of class c, and $\mu \in [0,1]$. We select $\mu$ such that the imbalance ratio---which is defined as the ratio between the number of examples in the largest and smallest class---is $50\times$ to $1000\times$.

\smallskip
\noindent \textbf{Sub-network training strategies.} We train a network on balanced Cifar-10/100 for 200 epochs with an initial learning rate of $0.1$ decayed by $0.01$ at epochs $160$ and $180$ using the cross entropy loss. On imbalanced versions, we introduce effective re-weighting ~\cite{cui2019class} at epoch $160$ and refer to this strategy as delayed re-weighting or DRW-160~\cite{cao2019learning}. 

\smallskip 
\noindent \textbf{Neural Architecture Search} We train a super-network for $600$ epochs with an initial learning rate of 0.1, decayed by 0.01 at epochs $400$ and $500$. On imbalanced datasets, re-weighting is applied at epoch 400. For searching the best subnet, we follow~\cite{guo2020single} and use an evolutionary search with $20$ generations, population of $50$, crossover number $25$, mutation number $25$, mutate probability $0.1$ and top-k of $10$.

\smallskip 
\noindent \textbf{Adaptation Strategies} To adapt a super-network, we fine-tune it for $200$ epochs with an initial LR of 0.01, decayed by $0.01$ at epoch 100. In case of procedure P1, we introduce re-weighting at epoch 1. For P2, we delay the re-weighting to epoch 100. For P3, we follow the NAS strategy detailed above.


\begin{table}[]

\begin{subtable}{0.49\textwidth}
\centering
    \begin{tabular}{cccccc}
    \toprule
    &  Adp & \multicolumn{4}{c}{Imbalance Ratio}                                 \\ \cmidrule(l){3-6}
    &             & 50$\times$     & 100$\times$      & 200$\times$   & 400$\times$   \\ \midrule
    baseline & P0           & 45.80        & 40.83                    & 36.30         &       32.80        \\ \cdashline{1-6}[.7pt/1.5pt]\noalign{\vskip 0.15em}
    & P1           & \textbf{45.06}        & \better \textbf{41.93}                     & \better \textbf{36.76}         &     \better \textbf{33.70}         \\
    & P2           & 44.86        & \better 41.86                    & \better 36.70         &   \better   33.46         \\ \cdashline{1-6}[.7pt/1.5pt]\noalign{\vskip 0.15em}
    paragon & P3          & 45.93        & 41.53                    & 37.03         &       33.40        \\ 
    \bottomrule
    \end{tabular}
    \caption{Cifar10-1$\times \longrightarrow$ Cifar100-$\{50, 100, 200, 400\}\times$}
    \label{subtab:baseline_paragon_c10_to_c100}
    \vspace{2mm}
\end{subtable}
\begin{subtable}{0.49\textwidth}
\centering
    \begin{tabular}{cccccc}
    \toprule
    & Adp & \multicolumn{4}{c}{Imbalance Ratio} \\ \cmidrule(l){3-6}
    &                  & 100$\times$    & 200$\times$   & 400$\times$   & 800$\times$   \\ \midrule
    baseline & P0              & 75.96   & 68.96  & 63.26  & 56.90   \\ 
    \cdashline{1-6}[.7pt/1.5pt]\noalign{\vskip 0.15em}
    & P1             & \textbf{75.93}   & \better \textbf{69.70}  & \better \textbf{63.80}  & \better \textbf{58.23}  \\
    & P2             & 75.86   & \better 69.26  & \better 63.70  & \better 58.03  \\
    \cdashline{1-6}[.7pt/1.5pt]\noalign{\vskip 0.15em}
    paragon & P3              & 76.03   & 70.23  & 63.96  & 57.70   \\ 
    \bottomrule
    \end{tabular}
    \caption{Cifar100-1$\times  \longrightarrow$ Cifar10-$\{100, 200, 400, 800\}\times$}
    \label{subtab:baseline_paragon_c100_to_c10}
\end{subtable}
\caption{ \textbf{Comparing rank adaptation strategies.} Given a NAS super-net trained on $\mathcal{D}_s$, we adapt it to $\mathcal{D}_t$ and search the optimal sub-nets. These are retrained from scratch on $\mathcal{D}_t$ and the average validation accuracy is presented. Note that sub-nets obtained via P1/P2 outperform P0 for high imbalance ratios (shaded yellow) and typically P1 outperforms P2 (the winner is bolded). Results averaged over three seeds.}
\label{tab:baseline_paragon}
\end{table}

\subsection{Baseline and Paragon for IMB-NAS}
Given a NAS super-network trained on a source dataset $\mathcal{D}_s$, our goal is to efficiently adapt it to the target dataset $\mathcal{D}_t$ following which, the best sub-net is searched in the adapted super-net. 
Table \ref{subtab:baseline_paragon_c10_to_c100} illustrates the results for the case when $\mathcal{D}_s$ is Cifar10, and $\mathcal{D}_t$ is Cifar100 with varying levels of imbalance.
The first row (i.e. P0) refers to the case when the best sub-nets obtained on $\mathcal{D}_s$ are re-trained on $\mathcal{D}_t$. This serves as our lower bound or \textit{baseline}.
The last row (i.e. P3) refers to the case when the NAS super-net is trained on $\mathcal{D}_t$. This serves as the upper bound or the \textit{paragon} of accuracy.
Our two adaptation procedures (P1, P2) in the middle rows are highlighted yellow when they outperform the baseline, and the better among the two is bolded.

Observe from Tables \ref{subtab:baseline_paragon_c10_to_c100},\ref{subtab:baseline_paragon_c100_to_c10} that both adaptation procedures comprehensively outperform the baseline at higher levels of imbalance.
This means that the architectures searched on $\mathcal{D}_s$ can no longer be assumed as the optimal ones on imbalanced target datasets.
Interestingly, between P1 and P2, we find that P1 consistently outperforms P2. 
This is surprising since P2 also adapts the NAS backbone on the target data whereas P1 re-uses the backbone from the source dataset.
We hypothesize this occurs because, class imbalance is much larger an issue for searching the NAS backbone than the domain difference between Cifar10 and Cifar100.

Overall, we find that P1 and P2 achieve very close accuracy to the paragon (P3) while avoiding much of the compute burden of P3 as illustrated in the next section.

\subsection{Dissecting the performance adaptation}
In this section, we analyze different aspects of procedures P1-P3 by applying them to adapt a NAS super-net pre-trained on Cifar10-1x onto Cifar100-100x.

\begin{figure}[t!]
    \centering
    \includegraphics[width=0.4\textwidth]{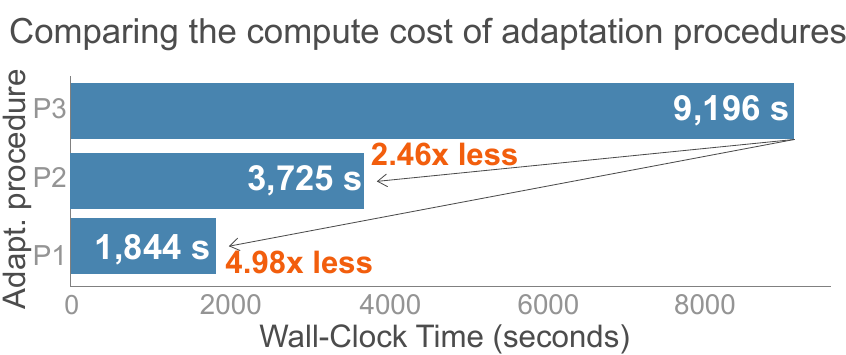}
    \caption{
     \textbf{Comparing the compute cost} of adapting a NAS super-net trained on Cifar10-$1\times$ onto Cifar100-$100\times$. The y-axis plots the wall clock time spent on a single V100 GPU. Observe that P2 and P3 consume $2\times$ and $5\times$ the cost of P1. Results averaged over three seeds.}
    \label{fig:adjustment_compute_cost}
\end{figure}

\smallskip \noindent
\textbf{Comparison on training cost.} We measure the wall-clock training time on a single V100 GPU as a proxy for training cost. 
The amortized training cost over three runs is presented in Fig~\ref{fig:adjustment_compute_cost}. 
It takes P1 $2000$ seconds to adapt a NAS super-network from cifar10-1$\times$ to cifar100-100$\times$.
In comparison P2 consumes $2\times$, and P3 consumes $5\times$ more time.
These results demonstrate that not only P1 can successfully adapt a super-net to improve accuracy, it is also very efficient.

\smallskip \noindent
\textbf{Impact of fine-tuning the backbone with P2.} In procedure P2, we adapt the NAS backbone via fine-tuning on the target dataset $\mathcal{D}_t$. 
One may wonder, can the backbone be frozen after the loss re-weighting is applied? 
The intuition being that re-weighting mainly helps adapt the classification boundary while negatively affecting the representation learned by the backbone~\cite{kang2019decoupling}.
To answer this, Table~\ref{tab:pt_ft_ablation} presents an ablation on the number of epochs spent on fine-tuning the NAS backbone with P2.
Observe that too few fine-tuning epochs (e.g. 50) leads to low sub-net accuracy. 
At the other end, fine-tuning for 100 epochs is sufficient to improve the sub-net accuracy beyond the paragon (P3).
This means that one could further lower the compute burden of P2 by freezing the backbone once loss re-weighting is applied at epoch 100.
\begin{table}[t!]
\centering
\begin{tabular}{ccccc}
\toprule
Proc                & Epochs & \multicolumn{3}{c}{Imbalance Ratio} \\ \cmidrule(l){3-5}
                    &    -    & 100$\times$       & 200$\times$       & 400$\times$      \\ \midrule
P0                  &    -    & 40.83      & 36.3       & 32.80     \\
\cdashline{1-5}[.7pt/1.5pt]\noalign{\vskip 0.15em}\
P1                  &   -     & 41.93      & 36.76      & 33.70     \\
\cdashline{1-5}[.7pt/1.5pt]\noalign{\vskip 0.15em}\
\multirow{4}{*}{P2} & 50     & 40.06      & 35.46      & 32.56     \\
                    & 100    & 41.43      & 37.26      & 33.00     \\
                    & 150    & 41.40       & 36.63      & 32.86     \\
                    & 200    & 41.86      & 36.70       & 33.46     \\
\cdashline{1-5}[.7pt/1.5pt]\noalign{\vskip 0.15em}
P3                  &    -    & 41.53      & 37.03      & 33.4      \\ \hline
\end{tabular}
\caption{\textbf{Ablating on} the number of backbone fine-tuning epochs with P2 while adapting from Cifar10-1$\times$ to Cifar100-$\{100,200,400\}\times$. Coinciding the freezing of the backbone with the loss re-weighting at epoch 100 typically outperforms the baseline. Generally more fine-tuning epochs are better. Results averaged over three seeds.}
\label{tab:pt_ft_ablation}
\end{table}

\begin{table}[t!]
\centering
\begin{tabular}{@{}lccccc@{}}
\toprule
Train Loss & \multicolumn{5}{c}{Imbalance Ratio}    \\ \cmidrule(l){2-4}
             & 100$\times$    & 200$\times$  & 400$\times$ \\ \midrule
CE         & 41.36   & 37.5 &  33.9   \\
CE-DRW     & 41.53   & 37.0 &  33.4   \\ \bottomrule
\end{tabular}
\caption{\textbf{Ablating on the loss} used to train the NAS super-net on Cifar100-100$\times$. Results presented are the validation accuracy of optimal sub-networks searched from corresponding super-networks. It is inconclusive if training the NAS super-net with re-weighted loss (CE-DRW) induces better sub-networks. Results averaged over three seeds.}
\label{tab:nas_training_ablation}
\end{table}

\smallskip \noindent
\textbf{Training the NAS super-net with loss re-weighting.} We observe that loss re-weighting generally results in improved super-net accuracy on imbalanced datasets. 
Does this mean the resulting sub-nets are better than the ones obtained from a super-net trained without loss re-weighting?
We answer this question we train super-nets on Cifar100-100x with and without re-weighting.
Then we search and train the best sub-nets which are presented in Table.~\ref{tab:nas_training_ablation}.
We find that there is no clear winner among the two NAS training approaches.

\begin{table}[t!]
\centering
\begin{adjustbox}{width=0.95\columnwidth}
\begin{tabular}{@{}ccccccccc@{}}
\toprule
Adj & \multicolumn{8}{c}{Imbalance Ratio}                         \\ 
    & \multicolumn{4}{c}{100$\times$}  & \multicolumn{4}{c}{400$\times$}  \\
 \cmidrule(l){2-5} \cmidrule(l){6-9}
    & High & Med  & Low  & All  & High & Med  & Low  & All  \\ \midrule
P0  & 64.1 & 40.4 & 14.1 & 40.8 & 65.0 & 39.1 & 10.1 & 32.8 \\
\cdashline{1-9}[.7pt/1.5pt]\noalign{\vskip 0.15em}
P1  & 6\better 5.2 & \better 41.6 & \better 15.0 & \better 41.9 & \better 66.3 & \better 41.0 & \better 10.2 & \better 33.7 \\
P2  & \better 65.2 & \better 40.9 & \better 15.7 & \better 41.8 & \better 66.2 & \better 40.3 & \better 10.2 & \better 33.4 \\
\cdashline{1-9}[.7pt/1.5pt]\noalign{\vskip 0.15em}

P3  & 65.0 & 41.5 & 14.1 & 41.5 & 66.2 & 39.5 & 10.5 & 33.4 \\ \bottomrule
\end{tabular}
\end{adjustbox}
\caption{\textbf{Dissecting the overall accuracy} on Cifar100-$\{100,400\}\times$ into the accuracy on classes containing many (i.e. $>100$), medium (i.e. between 20-100) and few (i.e. $<20$) examples per class. Sub-networks obtained via P1 and P2 outperform the baseline (P0) for all class categories (shaded yellow). Results averaged over three seeds. }
\label{tab:high_med_low_comparison}
\end{table}

\smallskip \noindent
\textbf{Dissecting the overall accuracy improvement.} To analyze which classes contribute to an increase accuracy, Table.~\ref{tab:high_med_low_comparison} dissects the overall accuracy (denoted by column ``All'')  into the accuracy obtained on classes containing Many (i.e. $>100$), Medium (i.e. between 20-100) and Few (i.e. $<20$) examples per class.
For both $100\times$ and $400\times$ levels of imbalance, the architectures obtained via P1 and P2 outperform those obtained by $P0$ for all class categories.
This means that indeed the architectures obtained via P1,P2 are able to learn better representations.


\section{Conclusion}
This work aims to improve performance on class imbalanced datasets by optimising the backbone architecture.
Towards this goal, we discover that an architecture's performance on balanced datasets is not indicative if its performance on imbalanced ones.
This observation suggests re-running NAS on each target dataset.
To overcome the prohibitive compute burden or re-running NAS, we propose to adapt a NAS super-net trained on balanced datasets onto imbalanced ones.
We develop multiple adaptation procedures and find that re-training the linear classification head while freezing the NAS super-net backbone outperforms other adaptation strategies both in terms of efficiency of the adaptation and the accuracy of the resulting sub-networks.

\bibliography{aaai22}

\begin{thebibliography}{31}
\providecommand{\natexlab}[1]{#1}

\bibitem[{Baker et~al.(2016)Baker, Gupta, Naik, and
  Raskar}]{baker2016designing}
Baker, B.; Gupta, O.; Naik, N.; and Raskar, R. 2016.
\newblock Designing neural network architectures using reinforcement learning.
\newblock \emph{arXiv preprint arXiv:1611.02167}.

\bibitem[{Baker et~al.(2017)Baker, Gupta, Raskar, and
  Naik}]{baker2017accelerating}
Baker, B.; Gupta, O.; Raskar, R.; and Naik, N. 2017.
\newblock Accelerating neural architecture search using performance prediction.
\newblock \emph{arXiv preprint arXiv:1705.10823}.

\bibitem[{Cao et~al.(2019)Cao, Wei, Gaidon, Arechiga, and Ma}]{cao2019learning}
Cao, K.; Wei, C.; Gaidon, A.; Arechiga, N.; and Ma, T. 2019.
\newblock Learning Imbalanced Datasets with Label-Distribution-Aware Margin
  Loss.
\newblock In \emph{Advances in Neural Information Processing Systems}.

\bibitem[{Chawla et~al.(2002)Chawla, Bowyer, Hall, and
  Kegelmeyer}]{chawla2002smote}
Chawla, N.~V.; Bowyer, K.~W.; Hall, L.~O.; and Kegelmeyer, W.~P. 2002.
\newblock SMOTE: synthetic minority over-sampling technique.
\newblock \emph{Journal of artificial intelligence research}, 16: 321--357.

\bibitem[{Chu et~al.(2020)Chu, Bian, Liu, and Ling}]{chu2020feature}
Chu, P.; Bian, X.; Liu, S.; and Ling, H. 2020.
\newblock Feature space augmentation for long-tailed data.
\newblock In \emph{European Conference on Computer Vision}, 694--710. Springer.

\bibitem[{Cui et~al.(2019)Cui, Jia, Lin, Song, and Belongie}]{cui2019class}
Cui, Y.; Jia, M.; Lin, T.-Y.; Song, Y.; and Belongie, S. 2019.
\newblock Class-balanced loss based on effective number of samples.
\newblock In \emph{Proceedings of the IEEE/CVF conference on computer vision
  and pattern recognition}, 9268--9277.

\bibitem[{Dong et~al.(2021)Dong, Liu, Musial, and Gabrys}]{dong2021nats}
Dong, X.; Liu, L.; Musial, K.; and Gabrys, B. 2021.
\newblock Nats-bench: Benchmarking nas algorithms for architecture topology and
  size.
\newblock \emph{IEEE transactions on pattern analysis and machine
  intelligence}.

\bibitem[{Duggal et~al.(2020)Duggal, Freitas, Dhamnani, Chau, and
  Sun}]{duggal2020elf}
Duggal, R.; Freitas, S.; Dhamnani, S.; Chau, D.~H.; and Sun, J. 2020.
\newblock Elf: An early-exiting framework for long-tailed classification.
\newblock \emph{arXiv preprint arXiv:2006.11979}.

\bibitem[{Duggal et~al.(2021{\natexlab{a}})Duggal, Freitas, Dhamnani, Chau, and
  Sun}]{duggal2021har}
Duggal, R.; Freitas, S.; Dhamnani, S.; Chau, D.~H.; and Sun, J.
  2021{\natexlab{a}}.
\newblock HAR: Hardness Aware Reweighting for Imbalanced Datasets.
\newblock In \emph{2021 IEEE International Conference on Big Data (Big Data)},
  735--745. IEEE.

\bibitem[{Duggal et~al.(2021{\natexlab{b}})Duggal, Zhou, Yang, Xiong, Xia, Tu,
  and Soatto}]{duggal2021compatibility}
Duggal, R.; Zhou, H.; Yang, S.; Xiong, Y.; Xia, W.; Tu, Z.; and Soatto, S.
  2021{\natexlab{b}}.
\newblock Compatibility-aware heterogeneous visual search.
\newblock In \emph{Proceedings of the IEEE/CVF Conference on Computer Vision
  and Pattern Recognition}, 10723--10732.

\bibitem[{Elsken, Metzen, and Hutter(2019)}]{elsken2019neural}
Elsken, T.; Metzen, J.~H.; and Hutter, F. 2019.
\newblock Neural architecture search: A survey.
\newblock \emph{The Journal of Machine Learning Research}, 20(1): 1997--2017.

\bibitem[{Falkner, Klein, and Hutter(2018)}]{falkner2018bohb}
Falkner, S.; Klein, A.; and Hutter, F. 2018.
\newblock BOHB: Robust and efficient hyperparameter optimization at scale.
\newblock In \emph{International Conference on Machine Learning}, 1437--1446.
  PMLR.

\bibitem[{Guo et~al.(2020)Guo, Zhang, Mu, Heng, Liu, Wei, and
  Sun}]{guo2020single}
Guo, Z.; Zhang, X.; Mu, H.; Heng, W.; Liu, Z.; Wei, Y.; and Sun, J. 2020.
\newblock Single path one-shot neural architecture search with uniform
  sampling.
\newblock In \emph{European conference on computer vision}, 544--560. Springer.

\bibitem[{He et~al.(2008)He, Bai, Garcia, and Li}]{he2008adasyn}
He, H.; Bai, Y.; Garcia, E.~A.; and Li, S. 2008.
\newblock ADASYN: Adaptive synthetic sampling approach for imbalanced learning.
\newblock In \emph{2008 IEEE international joint conference on neural networks
  (IEEE world congress on computational intelligence)}, 1322--1328. IEEE.

\bibitem[{Howard et~al.(2017)Howard, Zhu, Chen, Kalenichenko, Wang, Weyand,
  Andreetto, and Adam}]{howard2017mobilenets}
Howard, A.~G.; Zhu, M.; Chen, B.; Kalenichenko, D.; Wang, W.; Weyand, T.;
  Andreetto, M.; and Adam, H. 2017.
\newblock Mobilenets: Efficient convolutional neural networks for mobile vision
  applications.
\newblock \emph{arXiv preprint arXiv:1704.04861}.

\bibitem[{Kang et~al.(2019)Kang, Xie, Rohrbach, Yan, Gordo, Feng, and
  Kalantidis}]{kang2019decoupling}
Kang, B.; Xie, S.; Rohrbach, M.; Yan, Z.; Gordo, A.; Feng, J.; and Kalantidis,
  Y. 2019.
\newblock Decoupling representation and classifier for long-tailed recognition.
\newblock \emph{arXiv preprint arXiv:1910.09217}.

\bibitem[{Liu, Simonyan, and Yang(2018)}]{liu2018darts}
Liu, H.; Simonyan, K.; and Yang, Y. 2018.
\newblock Darts: Differentiable architecture search.
\newblock \emph{arXiv preprint arXiv:1806.09055}.

\bibitem[{Liu et~al.(2019)Liu, Miao, Zhan, Wang, Gong, and Yu}]{liu2019large}
Liu, Z.; Miao, Z.; Zhan, X.; Wang, J.; Gong, B.; and Yu, S.~X. 2019.
\newblock Large-scale long-tailed recognition in an open world.
\newblock In \emph{Proceedings of the IEEE/CVF Conference on Computer Vision
  and Pattern Recognition}, 2537--2546.

\bibitem[{Lu et~al.(2021)Lu, Sreekumar, Goodman, Banzhaf, Deb, and
  Boddeti}]{lu2021neural}
Lu, Z.; Sreekumar, G.; Goodman, E.; Banzhaf, W.; Deb, K.; and Boddeti, V.~N.
  2021.
\newblock Neural architecture transfer.
\newblock \emph{IEEE Transactions on Pattern Analysis and Machine
  Intelligence}, 43(9): 2971--2989.

\bibitem[{Menon et~al.(2020)Menon, Jayasumana, Rawat, Jain, Veit, and
  Kumar}]{menon2020long}
Menon, A.~K.; Jayasumana, S.; Rawat, A.~S.; Jain, H.; Veit, A.; and Kumar, S.
  2020.
\newblock Long-tail learning via logit adjustment.
\newblock \emph{arXiv preprint arXiv:2007.07314}.

\bibitem[{Panda et~al.(2021)Panda, Merler, Jaiswal, Wu, Ramakrishnan, Finkler,
  Chen, Cho, Feris, Kung et~al.}]{panda2021nastransfer}
Panda, R.; Merler, M.; Jaiswal, M.~S.; Wu, H.; Ramakrishnan, K.; Finkler, U.;
  Chen, C.-F.~R.; Cho, M.; Feris, R.; Kung, D.; et~al. 2021.
\newblock Nastransfer: Analyzing architecture transferability in large scale
  neural architecture search.
\newblock In \emph{Proceedings of the AAAI Conference on Artificial
  Intelligence}, volume~35, 9294--9302.

\bibitem[{Real et~al.(2017)Real, Moore, Selle, Saxena, Suematsu, Tan, Le, and
  Kurakin}]{real2017large}
Real, E.; Moore, S.; Selle, A.; Saxena, S.; Suematsu, Y.~L.; Tan, J.; Le,
  Q.~V.; and Kurakin, A. 2017.
\newblock Large-scale evolution of image classifiers.
\newblock In \emph{International Conference on Machine Learning}, 2902--2911.
  PMLR.

\bibitem[{Tian et~al.(2020)Tian, Liu, Glaser, Hsu, and
  Kira}]{tian2020posterior}
Tian, J.; Liu, Y.-C.; Glaser, N.; Hsu, Y.-C.; and Kira, Z. 2020.
\newblock Posterior re-calibration for imbalanced datasets.
\newblock \emph{Advances in Neural Information Processing Systems}, 33:
  8101--8113.

\bibitem[{Wang, Ramanan, and Hebert(2017)}]{wang2017learning}
Wang, Y.-X.; Ramanan, D.; and Hebert, M. 2017.
\newblock Learning to model the tail.
\newblock \emph{Advances in neural information processing systems}, 30.

\bibitem[{Wu et~al.(2020)Wu, Morgado, Wang, Ho, and
  Vasconcelos}]{wu2020solving}
Wu, T.-Y.; Morgado, P.; Wang, P.; Ho, C.-H.; and Vasconcelos, N. 2020.
\newblock Solving long-tailed recognition with deep realistic taxonomic
  classifier.
\newblock In \emph{European Conference on Computer Vision}, 171--189. Springer.

\bibitem[{Yin et~al.(2019)Yin, Yu, Sohn, Liu, and Chandraker}]{yin2019feature}
Yin, X.; Yu, X.; Sohn, K.; Liu, X.; and Chandraker, M. 2019.
\newblock Feature transfer learning for face recognition with under-represented
  data.
\newblock In \emph{Proceedings of the IEEE/CVF conference on computer vision
  and pattern recognition}, 5704--5713.

\bibitem[{Zhang et~al.(2021{\natexlab{a}})Zhang, Li, Yan, He, and
  Sun}]{zhang2021distribution}
Zhang, S.; Li, Z.; Yan, S.; He, X.; and Sun, J. 2021{\natexlab{a}}.
\newblock Distribution alignment: A unified framework for long-tail visual
  recognition.
\newblock In \emph{Proceedings of the IEEE/CVF conference on computer vision
  and pattern recognition}, 2361--2370.

\bibitem[{Zhang et~al.(2018)Zhang, Zhou, Lin, and Sun}]{zhang2018shufflenet}
Zhang, X.; Zhou, X.; Lin, M.; and Sun, J. 2018.
\newblock Shufflenet: An extremely efficient convolutional neural network for
  mobile devices.
\newblock In \emph{Proceedings of the IEEE conference on computer vision and
  pattern recognition}, 6848--6856.

\bibitem[{Zhang et~al.(2021{\natexlab{b}})Zhang, Kang, Hooi, Yan, and
  Feng}]{zhang2021deep}
Zhang, Y.; Kang, B.; Hooi, B.; Yan, S.; and Feng, J. 2021{\natexlab{b}}.
\newblock Deep long-tailed learning: A survey.
\newblock \emph{arXiv preprint arXiv:2110.04596}.

\bibitem[{Zhou et~al.(2020)Zhou, Cui, Wei, and Chen}]{zhou2020bbn}
Zhou, B.; Cui, Q.; Wei, X.-S.; and Chen, Z.-M. 2020.
\newblock Bbn: Bilateral-branch network with cumulative learning for
  long-tailed visual recognition.
\newblock In \emph{Proceedings of the IEEE/CVF conference on computer vision
  and pattern recognition}, 9719--9728.

\bibitem[{Zoph et~al.(2018)Zoph, Vasudevan, Shlens, and Le}]{zoph2018learning}
Zoph, B.; Vasudevan, V.; Shlens, J.; and Le, Q.~V. 2018.
\newblock Learning transferable architectures for scalable image recognition.
\newblock In \emph{Proceedings of the IEEE conference on computer vision and
  pattern recognition}, 8697--8710.

\end{thebibliography}




\end{document}